# High-Quality and Full Bandwidth Seismic Signal Synthesis using Operational GANs

Ozer Can Devecioglu, Serkan Kiranyaz, Zafer Yilmaz, Onur Avci, Moncef Gabbouj, and Ertugrul Taciroglu

*Abstract*—**Vibration sensors are essential in acquiring seismic activity for an accurate earthquake assessment. The state-of-the-art sensors can provide the best signal quality and the highest bandwidth; however, their high cost usually hinders a wide range of applicability and coverage, which is otherwise possible with their basic and cheap counterparts. But, their poor quality and low bandwidth can significantly degrade the signal fidelity and result in an imprecise analysis. To address these drawbacks, in this study, we propose a novel, high-quality, and full bandwidth seismic signal synthesis by transforming the signal acquired from an inferior sensor. We employ 1D Operational Generative Adversarial Networks (Op-GANs) with novel loss functions to achieve this. Therefore, the study's key contributions include releasing a new dataset, addressing operational constraints in seismic monitoring, and pioneering a deep-learning transformation technique to create the first virtual seismic sensor. The proposed method is extensively evaluated over the *Simulated Ground Motion* (SimGM) benchmark dataset, and the results demonstrated that the proposed approach significantly improves the quality and bandwidth of seismic signals acquired from a variety of sensors, including a cheap seismic sensor, the CSN-Phidgets, and the integrated accelerometers of an Android, and iOS phone, to the same level as the state-of-the-art sensor (e.g., Kinemetrics-Episensor). The SimGM dataset, our results, and the optimized PyTorch implementation of the proposed approach are publicly shared.**

## I. INTRODUCTION

The seismic waves are mechanical waves that pass through the ground. They are caused by earthquakes, volcanic eruptions, magma flows, or large human-made explosions that produce low-frequency waves. They propagate through the Earth's crust and are recorded, typically in the form of accelerations, either at the ground surface or at depth. Capturing these seismic signals at high spatiotemporal resolution is an essential capability in seismology and earthquake engineering in order to, for example, estimate the internal structure of the planet, track tectonic plate movements, and assess earthquake damage risks. In the literature, countless studies have been conducted to determine key characteristics of seismic signals, such as frequency content, spectral magnitude, peak ground acceleration, and energy release rates. These studies have focused on a wide variety of applications in examining, for example, volcanic activities [1], landslides [2], snow avalanches [3], rockslides [4], rock falls [5], earthquakes [6]-[9], and structural health and performance [10]-[12].

Several different types of vibration sensors are used for recording seismic signals. As in many other sensor types, vibration sensors in the market vary significantly in price and quality. Using relatively cheap sensors is attractive for end users who wish to increase the spatial density of records. Yet, such sensors invariably have limited capabilities, which can pose significant challenges in the ensuing analyses [13]. Cheap sensors often lack the necessary bandwidth, dynamic range, sensitivity, and noise-robustness required to accurately capture the inherently more frequent small and medium events or strong (rare) distant events [14], [15]. Both scenarios are far more common than recording strong nearby earthquakes at high spatial resolution and, as such, offer the best observational opportunities for advancing the state-of-the-art in seismology and earthquake engineering fields. Therefore, improving the capabilities of cheaper sensors can make a large impact [16].

As a typical example, *Figure 1* displays the simultaneous temporal and spectral representations of a seismic signal recording from the SimGM dataset using CSN-Phidgets (CSN), Kinemetrics-Episensor (Episensor), and the integrated accelerometers of the Xiaomi-Poco X3 Pro (Android), and iPhone 7 (iOS) devices. The Episensor and CSN sensors are specialized vibration sensors designed for professional seismic monitoring. Episensors represent the high-end devices in the market and offer low-noise measurements within a broad bandwidth. They can be considered as a benchmark device for earthquake research and monitoring. However, an Episensor typically comes with a higher price, requiring users to resort to lower-tier sensors, especially when high spatial density is required. As an alternative, CSN sensors are cost-effective but offer a limited bandwidth and sensitivity. Accelerometers and

O. Devecioglu and M. Gabbouj are with the Department of Computing Sciences, Tampere University, Tampere, Finland (e-mail: ozer.devecioglu@tuni.fi, moncef.gabbouj@tuni.fi).
S. Kiranyaz is with the Electrical Engineering Department, Qatar University, Doha, Qatar (e-mail: mkiranyaz@qu.edu.qa

Z. Yilmaz and E. Taciroglu is with the Department of Civil and Environmental Engineering, University of California, Los Angeles, CA, USA. (email: zafer@ucla.edu , etacir@ucla.edu )
O. Avci is with the Department of Civil and Environmental Engineering, West Virginia University, Morgantown, WV, USA (email: onur.avci@mail.wvu.edu ) gyroscopes built into Android and iOS phones can also record mechanical vibrations. However, they are not optimized for seismic monitoring, further decreasing accuracy and sensitivity. *Figure 1* reveals that the signals acquired by these sensors have different bandwidths and sampling frequencies. Specifically, the bandwidths for the Android sensor, iPhone sensor, CSN sensor, and Episensor are 100 Hz, 50 Hz, 25 Hz, and 100 Hz, respectively. As a result, neither CSN nor iPhone sensors can capture high-frequency seismic events. Android sensors, on the other hand, while having the same bandwidth, induce a high level of noise and some false spectral peaks, which degrades the signal fidelity.

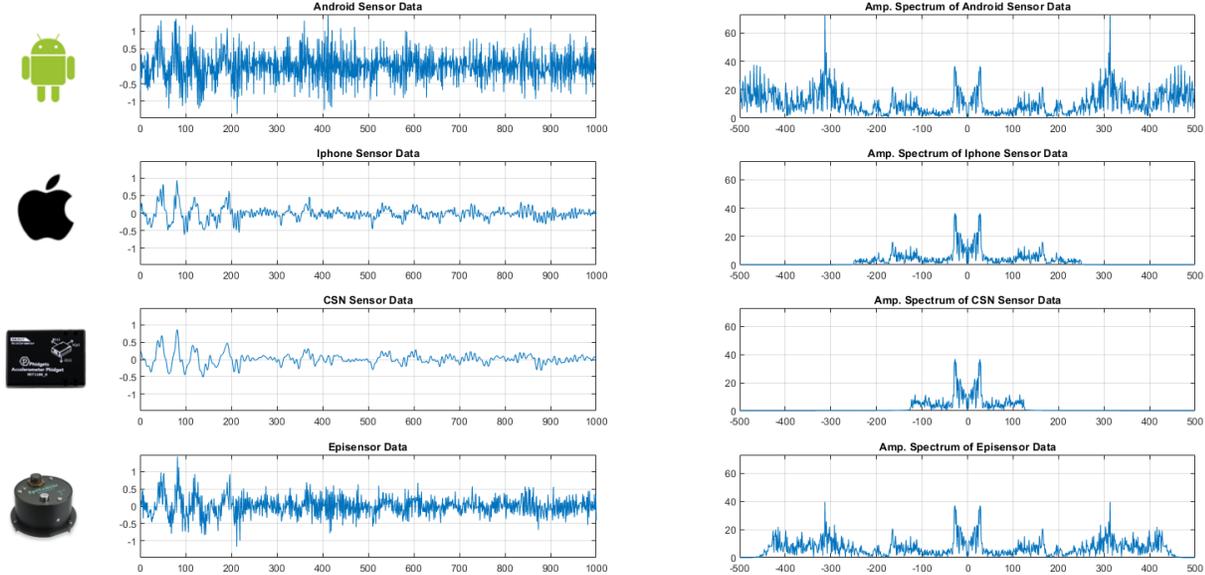

*Figure 1. Seismic signals in temporal (left column) and spectral (right column) domains simultaneously acquired by the Android (top), iPhone (2ⁿᵈ row), CSN (3ʳᵈ row), and Episensor (bottom) sensors.*

To improve the seismic signal quality and bandwidth, we propose a novel transformation technique in this study. The ultimate objective is to obtain the superior signal quality and bandwidth of the state-of-the-art sensor, the Episensor. Alternatively, the proposed method aims to create a virtual Episensor from any inferior seismic sensor by signal transformation. To accomplish this, we use a modified 1-D Operational GAN (Op-GAN) model where the convolutional layers of the Conditional GANs [17] are replaced with Self-Organized Operational Neural Networks (Self-ONNs) [18]-[25] layers with novel loss functions. Self-ONNs [26]-[37] are heterogeneous networks that can be defined as the superset of CNNs with a configurable parameter, *Q*, that determines the degree of nonlinearity (the degree of the polynomials) of each kernel transformation. Thus, a Self-ONN reduces to a CNN when *Q*=1 for every neuron in the network. The network can generate the optimal basis functions for achieving the highest learning performance because of the nonlinear nodal operator's "on-the-fly" generation. This means that a Self-ONN can outperform an equivalent or even a much deeper and more complex CNN thanks to its optimized nonlinearity and heterogeneity.

The rest of this article is organized as follows: The proposed 1-D Op-GAN model is presented in Section II with a brief explanation of 1-D Self-ONNs. In Section III, the benchmark SimGM seismic dataset is introduced first. Then, the performance of the proposed model over the SimGM dataset is evaluated. Conclusions and topics for future research are presented in Section IV.

## II. METHODOLOGY

In this section, we first provide a brief overview of 1D Self-ONNs and their key characteristics. Next, we present the 1D Op-GAN model trained for high-quality and full bandwidth seismic signal synthesis.

### A. 1D Self-Organized Operational Neural Networks

In this section, we briefly explore Self-ONNs and some of their main properties. Unlike the convolution operator of CNNs, the nodal operator of each generative neuron of a Self-ONN can perform any nonlinear transformation that can be represented using Taylor approximation near origin:

$$\psi(x) = \sum_{n=0}^{\infty} \frac{\psi^{(n)}(0)}{n!} x^n \qquad (1)$$

The $Q^{th}$ order truncated approximation, formally known as the Taylor polynomial, is represented by the following finite summation:

$$\psi(x)^{(Q)} = \sum_{n=0}^{Q} \frac{\psi^{(n)}(0)}{n!} x^n \qquad (2)$$

The above formulation can approximate any arbitrary function $\psi(x)$ near 0. When the activation function bounds the neuron's input feature maps in the vicinity of 0 (e.g., *tanh*), the formulation in (2) can be exploited to form a composite nodal operator where the power coefficients, $\frac{\psi^{(n)}(0)}{n!}$, can be the parameters of the network learned during training.

It was shown in [27]-[29] that the 1D nodal operator of the $k^{th}$ generative neuron in the $l^{th}$ layer takes the following general form:

$$\widetilde{\psi_k^l}\left(w_{ik}^{l(Q)}(r), y_i^{l-1}(m+r)\right)$$
$$= \sum_{q=1}^{Q} w_{ik}^{l(Q)}(r,q)\left(y_i^{l-1}(m+r)\right)^q \quad (3)$$

Let $x_{ik}^l \in \mathbb{R}^M$ be the contribution of the $i^{th}$ neuron at the $(l-1)^{th}$ layer to the input map of the $l^{th}$ layer. Therefore, it can be expressed as,

$$\widetilde{x_{ik}^l}(m) = \sum_{r=0}^{K-1}\sum_{q=1}^{Q} w_{ik}^{l(Q)}(r,q)\left(y_i^{l-1}(m+r)\right)^q \quad (4)$$

where $y_i^{l-1} \in \mathbb{R}^M$ is the output map of the $i^{th}$ neuron at the $(l-1)^{th}$ layer, $w_{ik}^{l(Q)}$ is a learnable kernel of the network, which is a $K \times Q$ matrix, i.e., $w_{ik}^{l(Q)} \in \mathbb{R}^{K \times Q}$, formed as, $w_{ik}^{l(Q)}(r) = [w_{ik}^{l(Q)}(r,1), w_{ik}^{l(Q)}(r,2), ..., w_{ik}^{l(Q)}(Q)]$. By the commutativity of the summation operations in (4), one can alternatively express:

$$\widetilde{x_{ik}^l}(m) = \sum_{q=1}^{Q}\sum_{r=0}^{K-1} w_{ik}^{l(Q)}(r, q-1)y_i^{l-1}(m+r)^q \quad (5)$$

It can be simplified as follows:

$$\widetilde{x_{ik}^l} = \sum_{q=1}^{Q} Conv1D\left(w_{ik}^{l(Q)}, \left(y_i^{l-1}\right)^q\right) \quad (6)$$

Hence, the formulation can be accomplished by applying Q 1D convolution operations. Finally, the output of this neuron can be formulated as follows:

$$x_k^l = b_k^l + \sum_{i=0}^{N_{l-1}} x_{ik}^l \quad (7)$$

where $b_k^l$ is the bias associated with this neuron. The $0^{th}$ order term, $q = 0$, the DC bias, is ignored as its additive effect can be compensated by the learnable bias parameter of the neuron. With the $Q = 1$ setting, a *generative* neuron reduces back to a convolutional neuron.

The raw-vectorized formulations of the forward propagation and detailed formulations of the Back-Propagation (BP) training in the raw-vectorized form can be found in [18], [20], and, [27].

### B. Seismic Signal Synthesis using 1-D Operational GANs

The primary objective is to transform signals acquired from cheap or non-professional sensors to the corresponding signal from the Episensor. In other words, we aim to create a virtual Episensor from such inferior sensors. To accomplish this, a dedicated 1-D Op-GAN will be configured and trained. As shown in *Figure 2*, the proposed 1-D Op-GAN model consists of an Operational Generator (OG) and Operational Discriminator (OD). As in a typical conditional GAN, the main target of the OG is to synthesize the (fake) Episensor signal in the output layer using the signal from an inferior sensor as the input. On the other hand, the OD forces the OG model to synthesize more realistic signals by discriminating the generator-synthesized *fake* signal samples from the *real* (target) signals. After the training step, the OD will be discarded since OG is the sole transformer network needed to accomplish the objective. In this work, each signal segment duration is 5 seconds with the sampling frequency of 200 Hz. Therefore, 5x200=1000 samples are windowed from each corresponding sensor signal. CSN and iPhone segments are interpolated by 4 and 2 times in order to keep the same segment size. All segments are normalized between [-1, 1] by,

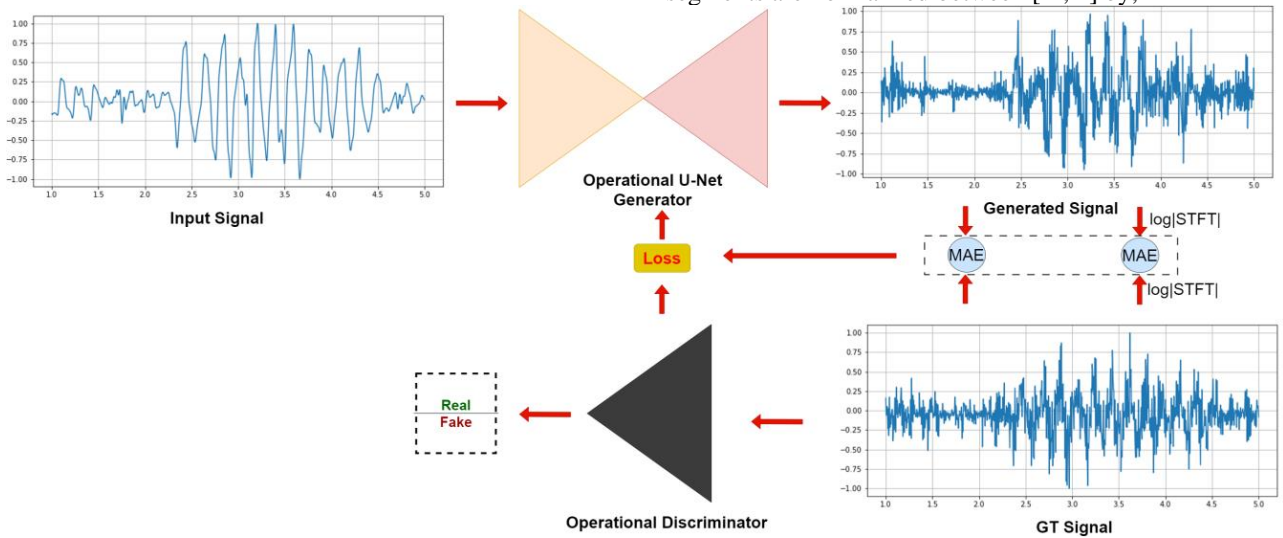

*Figure 2. The training scheme of the proposed 1D Op-GAN synthesis network.*

$$X_N(i) = \frac{2(X(i) - X_{\min})}{X_{max} - X_{min}} - 1 \qquad (8)$$

where $X(i)$ is the $i^{\text{th}}$ original sample amplitude in the segment, $X_N(i)$ is the $i^{\text{th}}$ sample amplitude of the normalized segment, $X_{min}$ and $X_{max}$ are the minimum and maximum amplitudes within the segment, respectively. This will scale the segment linearly in the range of [-1 1], where $X_{min} \rightarrow -1$ and $X_{max} \rightarrow 1$.

The objective function of 1D Op-GAN can be expressed as,

$$
\begin{aligned}
\min_{OG} \max_{OD} & L_{OP-GAN}(OG, OD) \\
= & E[\log(OD(GT))] \\
& + E[\log(1 - OD(OG(X)))]
\end{aligned}
\qquad (9)
$$

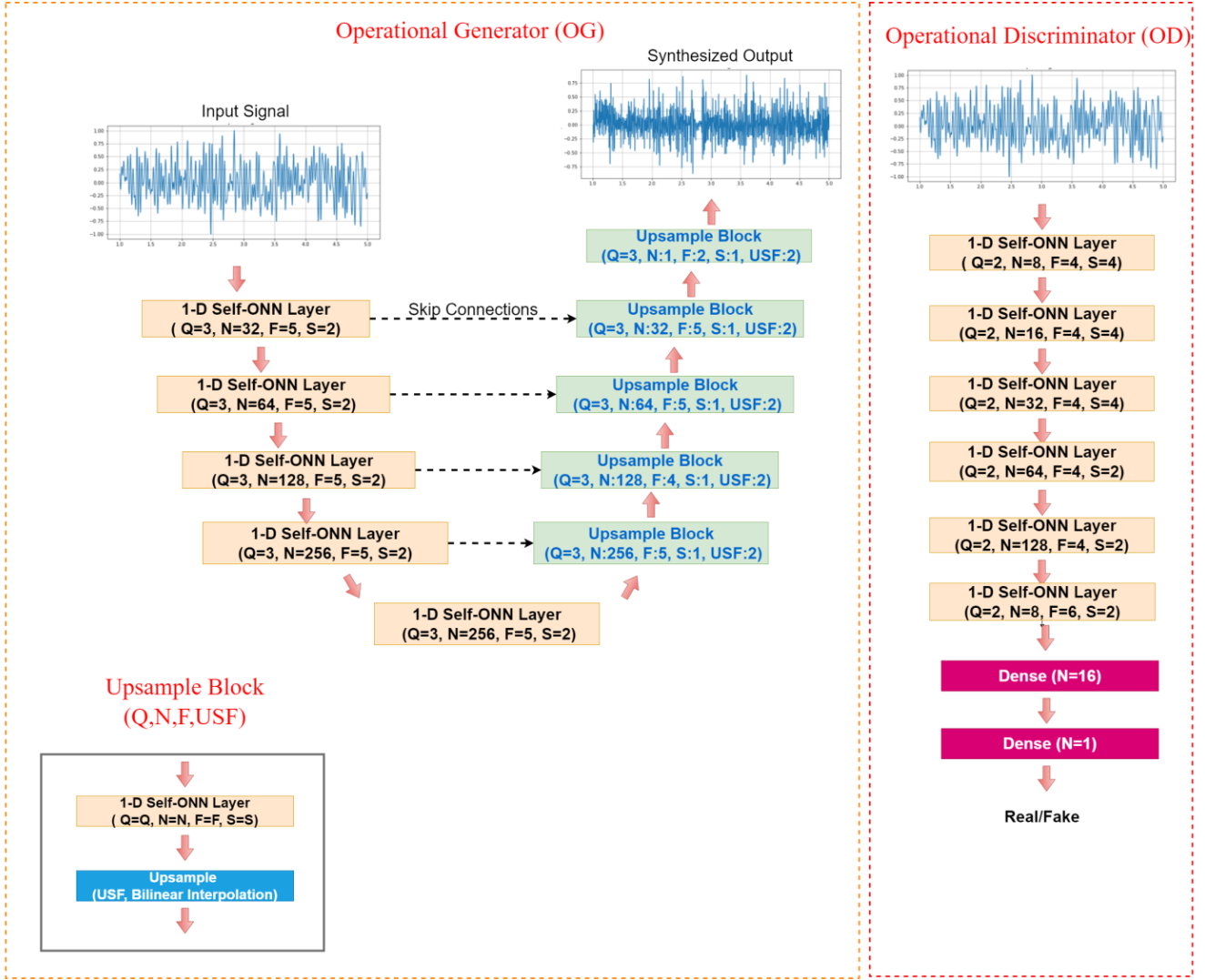

**Figure 3. The generator and discriminator architectures of the 1D Op-GAN**

On the adversarial loss function $L_{OP-GAN}(OG, OD)$, where X is the input (source) signal and GT is the corresponding ground truth (target) signal. The OG and OD compete in a two-player min-max game during the adversarial training. The discriminator's goal is to maximize $\log(OD(GT))$ and $\log(1 - OD(OG(X)))$ whereas the generator wants to reduce and $\log(1 - OD(OG(X)))$. However, unlike traditional GANs, training objective function of the proposed Op-GANs are modified with spectral loss functions. As mentioned earlier, there are significant disparities in the spectra of different sensor signals. It is clear that Episensor signals (GT), have the full bandwidth with the true spectral events (peaks in the frequency domain). In order to match these peaks in the frequency domain, we have penalized the total loss function in both spectral and temporal domains.

$$Loss_{Time} = \|(GT - OG(X))\|_1 \qquad (10)$$

Eq (10) expresses the Mean-Absolute Error (MAE) loss function in the time domain. In order to represent the spectrum, an N-point discrete STFT of the input and output signals is first computed. Eq. (11) expresses the complex-valued $N$-point discrete STFT from which the $N$-point discrete spectrogram, $Spec(X(n,k))$, can be computed as expressed in Eq. (12). We used *N*=256 samples long *Hanning* window with 128 samples overlap. Eq. (13) formulates the spectral loss function.

$$STFT[X, w, n] = X(n, w) = \sum_m^{\square} X[m]W[n - m]e^{-jwm} \quad (11)$$

$$X(n, k) = X(n, w)\big|_{w=\frac{2\pi k}{N}} \rightarrow Spec\big(X(n, k)\big) = |X(n, k)|^2 \quad (12)$$

$$Loss_{STFT} = \|STFT(GT) - Synth(X)_{\square}\|_1 \quad (13)$$

The objective of Operational-GAN training is to minimize the total loss in (14):

$$Loss_{total} = \lambda_1 L_{Op-GAN}(G) + \lambda_2 Loss_{Time} + \lambda_3 Loss_{STFT} \quad (14)$$

For the OG model, a 10-layer U-Net configuration is used with 5 1-D operational layers and 5 upsampling (by 2) and operational layers with residual connections (instead of transposed convolution layers). The kernel sizes are all set as 1x5. The stride is set as 2 for all down-sampling operational layers. For the encoder side of the OG, kernel sizes are set as 5, 4, 5, 5, and 2, respectively. The OD model consists of 6 operational layers with a kernel size of 4. The strides for layers are set as 4, 4, 4, 2, 2, and 2, respectively. As a loss function in the OD, mean absolute error (MAE) is computed between the discriminator output and label vectors. The architectures for the generators and discriminators are shown in Figure 3.

## III. EXPERIMENTAL RESULTS

### A. SimGM Benchmark Dataset and Experimental Setup

The ground motion acceleration data presents various valuable uses in areas such as seismic monitoring and early warning systems, structural health monitoring, and many fields of engineering. A total of 201 ground motions with varying moment magnitudes (Mw) and closest distance to the surface projection of rupture plane (Rjb), all recorded after 1997 and occurred on strike-slip fault type, was utilized to effectively test the capabilities of measurement instruments [37]. These ground motions are selected as two categories of Rjb (Joyner-Boore distance), one being considerably closer to the fault than the other, and eight categories of Mw with proportionally increasing released energy levels, as presented in *Table 1*. The distance to the epicenter can significantly affect the amplitude and duration of ground motion, while the ground properties between the epicenter and structure can modify the frequency content of the ground motion.

The ground motions are simulated on a displacement-controlled hydraulic pump-fed uni-directional shake table. Due to the displacement limit of the shake table being ± 1 inch, ground motion accelerations of high-magnitude earthquakes were slightly filtered until the criteria were met. Upon preparing the excitations, the tests were performed on the shake table where each sensor was placed.

*Table 1. The selected ground motions.*

| Magnitude (M_w) | Number of Ground Motions | |
|---|---|---|
| | 5-15 km R_{jb} | 90-110 km R_{jb} |
| 4.0-4.5 | 15 | 15 |
| 4.5-5.0 | 15 | 15 |
| 5.0-5.5 | 12 | 15 |
| 5.5-6.0 | 14 | 10 |
| 6.0-6.5 | 15 | 15 |
| 6.5-7.0 | 15 | 15 |
| 7.0-7.5 | 11 | 15 |
| 7.5-8.0 | 3 | 1 |

For all experiments, we employ a training scheme with a maximum of 5000 BP iterations and a batch size of 1. The Adam optimizer with the learning rates of $10^{-5}$ and $2.10^{-5}$ is used for the OG and the OD, respectively. The loss weights $\lambda_1$, $\lambda_2$, and $\lambda_3$ in (9) are set as 0.5, 0.1, and 0.1 during training. We implemented the proposed 1D Self ONN architectures using the FastONN library [20] based on Python and PyTorch and the final 1D Op-GAN is shared in [21].

*Table 2. Overall test performance of Op-GAN in terms of PSNR in temporal and spatial domains*

| Input Signal recorded by | | PSNR at Time Domain | PSNR at Frequency Domain |
|---|---|---|---|
| CSN Sensor | Input | 15.50 | 10.20 |
| | Output | 18.08 | 15.35 |
| Android Sensor | Input | 15.67 | 16.96 |
| | Output | 18.25 | 15.25 |
| iPhone Sensor | Input | 17.30 | 12.25 |
| | Output | 17.41 | 16.20 |

### B. Quantitative and Qualitative Evaluation

This section presents quantitative and qualitative (visual) evaluations of the proposed transformation technique. To compute the quantitative results, a common performance metric, the Peak-to-Signal-Noise-Ratio (PSNR) is utilized. The PSNR can be expressed as follows:

$$PSNR = 10 \log_{10}\left(\frac{(\max(Y_i))^2}{MSE(Y_i, GT_i)}\right) \quad (15)$$

where $Y_i$ is the output (synthesized) signal for the Op-GAN and $GT_i$ is the corresponding Ground Truth (target) signal.

Table *2* shows the temporal and spectral PSNR performances of the proposed transformation technique for all *sources*, i.e., CSN, Android, and iPhone, sensors with respect to the *target* sensor, the Episensor. The results indicate significant temporal/spectral PSNR gains. This is despite the fact that PSNR is not an adequate quantitative measure since it considers the overall (sample-by-sample) difference, not focusing on the major temporal and spectral events (e.g., peaks) in particular. Further analysis on the test results from CSN-recorded input signals reveals that 2.5dB and 5dB PSNR gains in the time and frequency domain, respectively, is achieved, which confirms the consistency of the signal synthesis. For iPhone Sensor with the half the bandwidth, around 4 dB average PSNR gain has been established. Finally, for the Android sensor, 2.5 dB PSNR gain was achieved; while there was a loss around 1.7 dB in the frequency domain. However, such a performance loss was not encountered on the qualitative evaluations where the transformed spectra can actually match most of the major spectral events of the target signal. Once again this is due to the fact that the PSNR is biased with the background (irrelevant) spectral events rather than the major ones.

For qualitative evaluation, Figure 4 illustrates three sets of input signals from three source sensors, the transformed and the target (GT) signals. Once again both temporal and spectral signal representations are presented. Additionally, on the top of each set, quantitative results for the respective sets are provided, along with corresponding temporal and spectral PSNR differences. 10 more results are presented in the Appendix section and 50 more [21]. All results indicate that the synthesized signals show a significant temporal and spectral similarity with the corresponding target signals, and in particular, the major spectral peaks of the target signal are synthesized successfully. As an example, the results shown in Figure 4 reveal that the transformed signals from the Android sensor, the target spectral peaks at around 30 Hz, 50 Hz, and 75Hz have been synthesized while suppressing other noise-like harmonics to align it more closely with the Episensor's spectrum. When we compare the synthesized Episensor signal with GT in the time domain, the similarity of the temporal patterns between the synthesized and target signals is quite high.

Especially, for the CSN signals as the input, 75% of the spectral elements (spectrum in the range of [25Hz, 100Hz]) are completely missing compared to the target (Episensor) signal. Despite this fact, the proposed transformation technique is able to synthesize those missing spectral events with elegant accuracy. Particularly the results indicate that most of the missing major spectral peaks in the frequencies higher than 25Hz are generated successfully from scratch. This is, indeed a more *synthesis* operation than a transformation. Moreover, in the example shown in Figure 4 (middle), during the synthesis of high-frequency spectrum, it adjusts the low-frequency part to surpass the false high peaks around 20 Hz. Finally, similar observations can be made for the iPhone signals as the input. This time half of the target bandwidth is missing (spectrum in the range of [50Hz, 100Hz]). Once again, the target spectral peaks higher than 50Hz are generated with an elegant accuracy and the spectral noise of the input signal has significantly been suppressed.

## C. Computational Complexity Analysis

The network size, total number of parameters (PARs), and inference time for each network configuration are computed in this section. Detailed formulations of the PARs calculations for Self-ONNs are available in [30]. A 2.2 GHz Intel Core i7 computer with 16 GB of RAM and an NVIDIA GeForce RTX 3080 graphics card was used for all experiments. The FastONN library [20] and Pytorch are used to implement the 1D Op-GAN network. The OG model has a total of 377K parameters. The processing time to synthesize a 5-second segment takes around 65msec for a single CPU implementation. This shows that the proposed transformation can achieve 75 times faster than the real-time requirements with a single CPU, indicating the potential of a real-time implementation even on low-cost, low-power sensorial hardware.

**Android-Recorded Signal**

Input PSNR: 13.0 | Output PSNR: 17.4 || FFT Input PSNR: 8.73 | FFT Output PSNR: 17.13

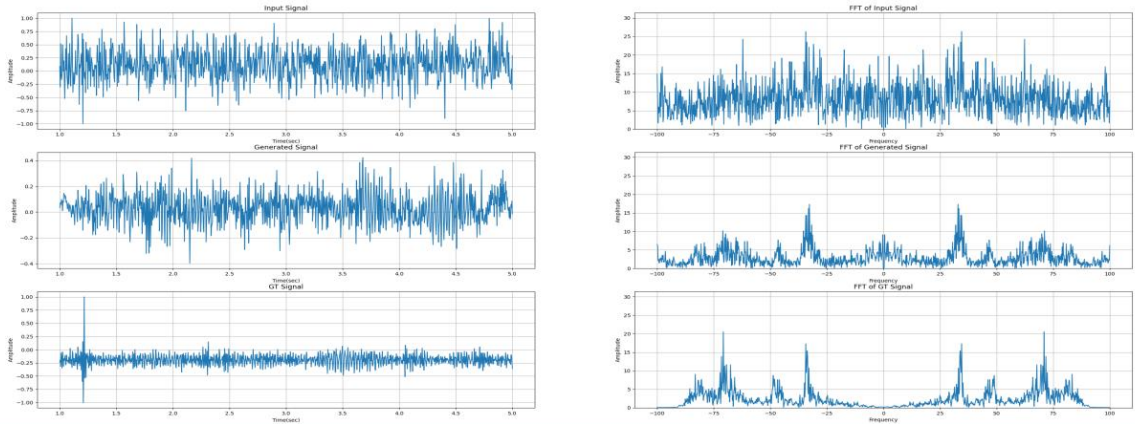

**CSN-Recorded Signal**

Input PSNR: 18.12 | Output PSNR: 20.48 || FFT Input PSNR: 11.81 | FFT Output PSNR: 18.4

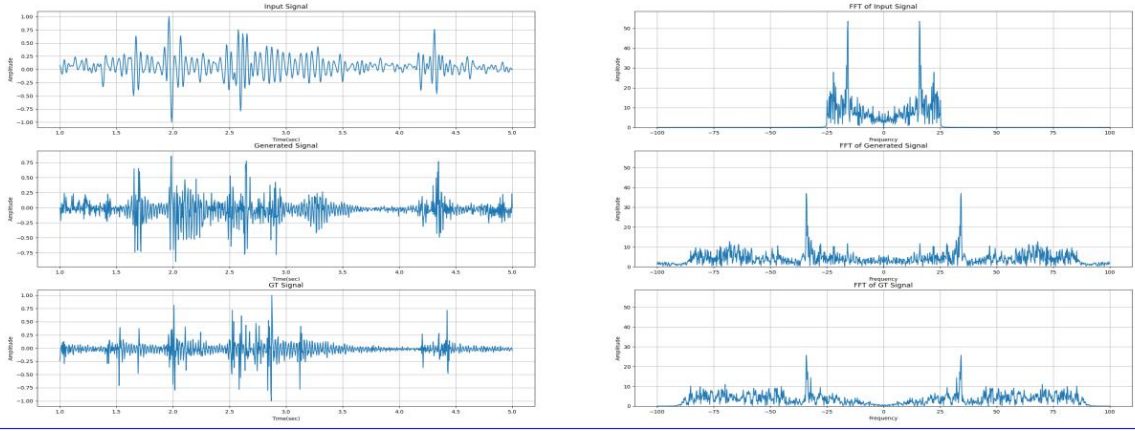

**iPhone-Recorded Signal**

Input PSNR: 16.35 | Output PSNR: 17.83 | FFT Input PSNR: 10.16 | FFT Output PSNR: 15.07

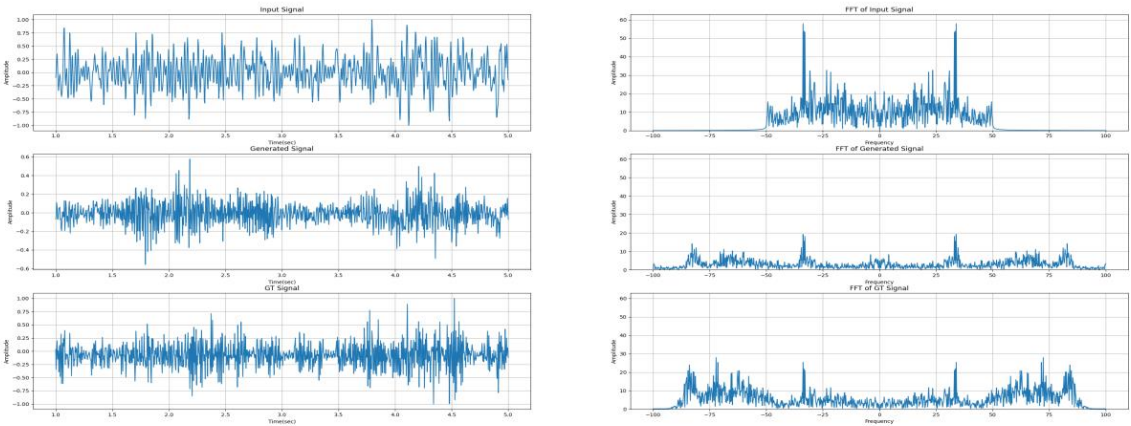

*Figure 4. The three sets of high quality and full bandwidth seismic signal synthesis results from corresponding Android, CSN, and iPhone signals, respectively. For each set, the first row shows the input (source) signal, the middle row is the transformed signal, and the bottom row shows the target (GT) signal from the Episensor.*

## IV. CONCLUSIONS

Vibration sensors play a crucial role in the analysis and identification of seismic signals. Using poor quality and cheap sensors can pose severe drawbacks and limitations for seismic monitoring. In this study, we propose a novel seismic signal synthesis model that can transform poor quality and low bandwidth signals acquired from such inferior sensors to a similar signal quality and bandwidth by the Episensor. The quantitative and qualitative results demonstrate that regardless of the input sensor and the signal quality, and bandwidth, the proposed method can significantly improve the signal quality in time and frequency domains. As a consequence, since the proposed approach does not require any specialized sensors, it significantly improves the practicality and accessibility of seismic monitoring. Additionally, it performs high efficiency, affordability, and resilience, as it can elevate the quality and extend the constrained bandwidth of sensors like CSN and iPhone. As a result, the proposed approach has the potential to revolutionize the field of seismic monitoring and make the process more accessible and practical for various other sensing applications.

The significant and novel contributions of this study can be summarized as follows:

- This study presents and publicly shares *SimGM Dataset*, which contains total of 26 hours of seismic signals acquired simultaneously from four different sensors.
- This pilot study demonstrates that poor-quality seismic signals possibly with a low bandwidth can be transformed to a similar level of a state-of-the-art sensor.
- Alternatively, this study demonstrates that a *virtual* state-of-the-art seismic sensor can be formed over any inferior sensor with the proposed transformation technique.
- Due to the elegant computational efficiency achieved using 1D Self-ONN OG model with a low depth and complexity, the proposed approach can be implemented in real-time even on low-cost, low-power sensorial hardware.

We can foresee that further improving the synthesis quality is possible by performing simultaneous recognition of the major temporal and spectral events on the seismic signal so that the transformation operation can be tuned to ignore the background noise-like signal. This will be the topic for our future research.

# APPENDIX

In this appendix, as shown in the plots from Figure 5 to Figure 14, we provide 10 additional sets of high-quality and full-bandwidth seismic signal synthesis results from the corresponding CSN, iPhone, and Android signals, respectively.

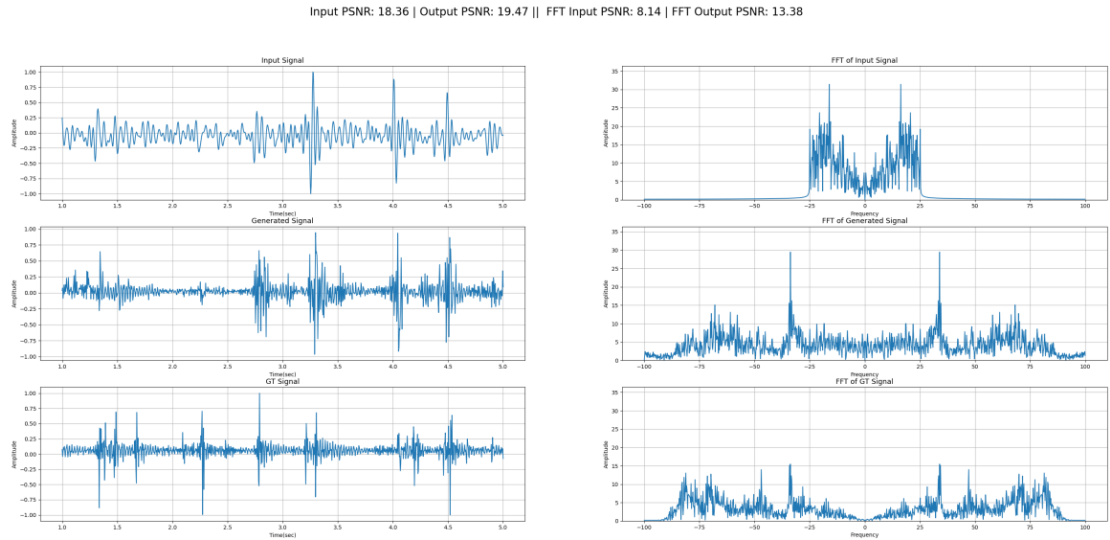

*Figure 5. High-quality and full-bandwidth seismic signal synthesis results from the corresponding CSN signal. The first row shows the input (source) signal, the middle row is the transformed signal, and the bottom row shows the target (GT) signal from the Episensor.*

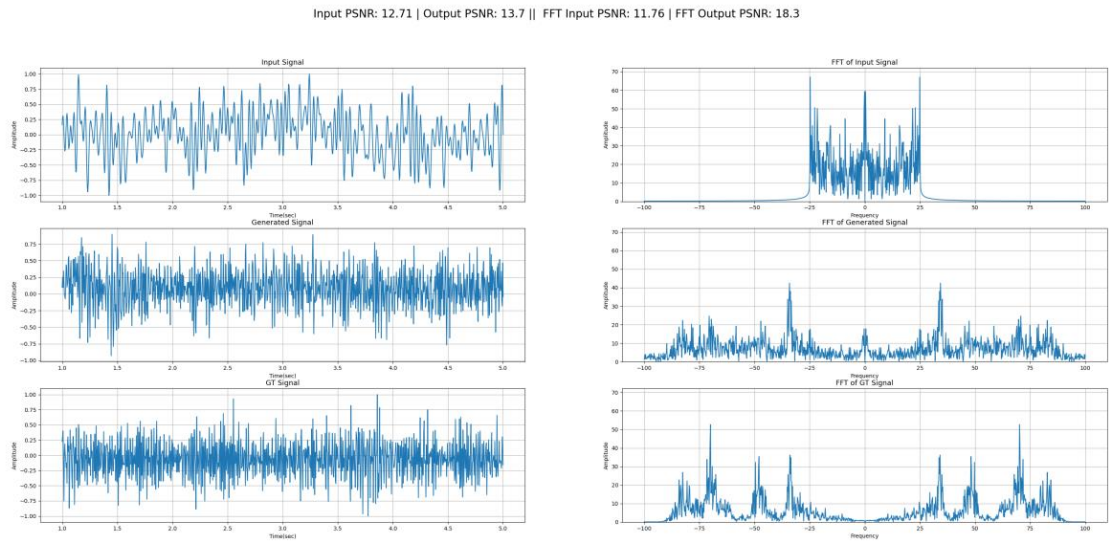

*Figure 6. High-quality and full-bandwidth seismic signal synthesis results from the corresponding CSN signal. The first row shows the input (source) signal, the middle row is the transformed signal, and the bottom row shows the target (GT) signal from the Episensor.*

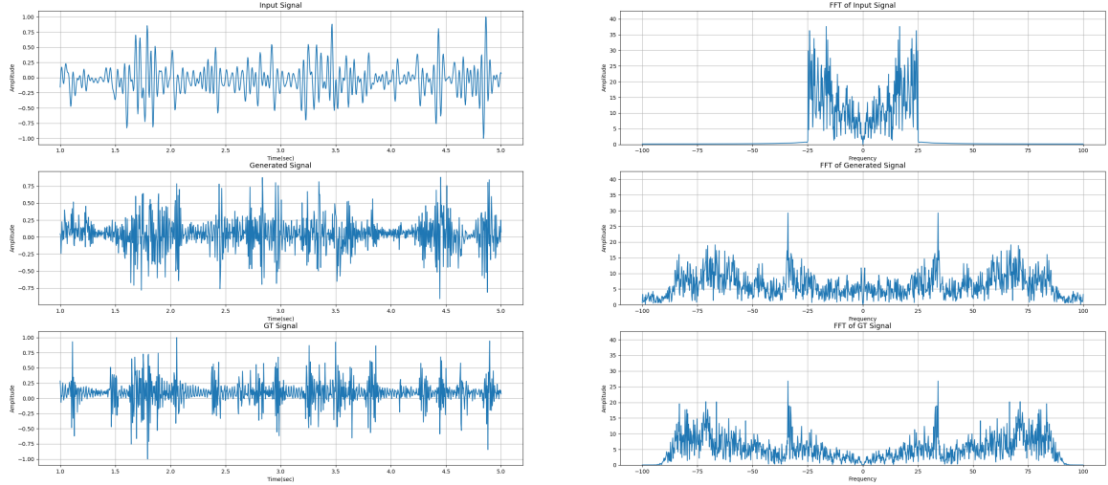

*Figure 7. High-quality and full-bandwidth seismic signal synthesis results from the corresponding CSN signal. The first row shows the input (source) signal, the middle row is the transformed signal, and the bottom row shows the target (GT) signal from the Episensor.*

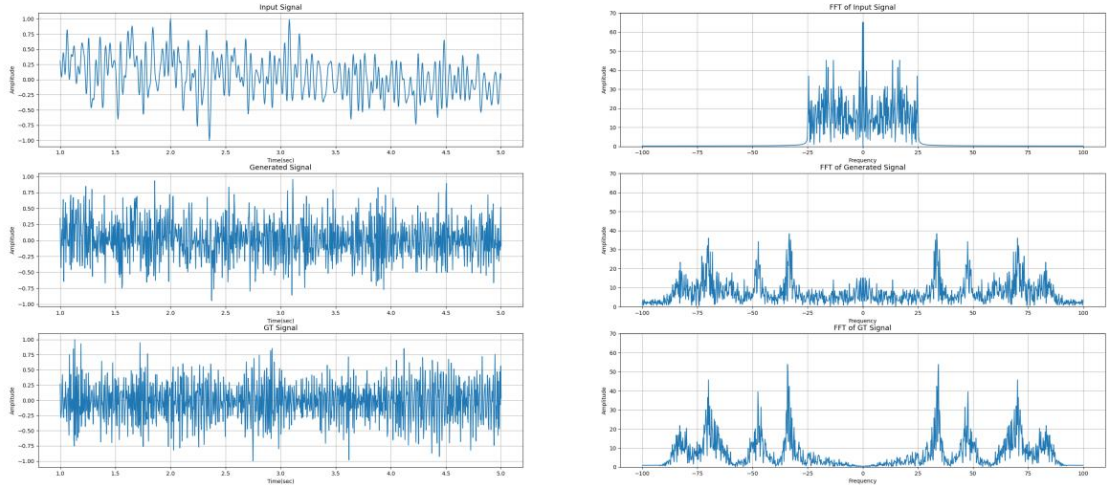

*Figure 8. High-quality and full-bandwidth seismic signal synthesis results from the corresponding CSN signal. The first row shows the input (source) signal, the middle row is the transformed signal, and the bottom row shows the target (GT) signal from the Episensor.*

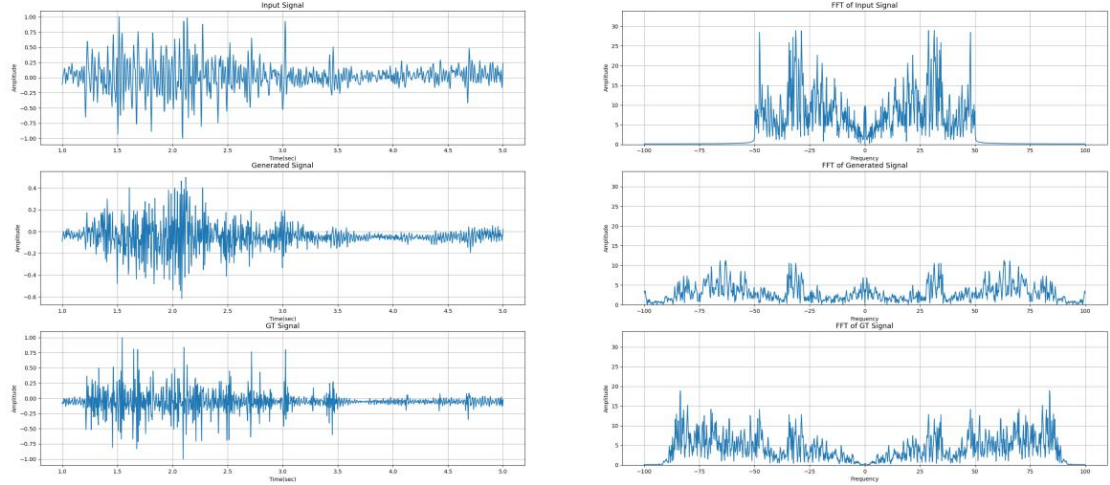

*Figure 9. High-quality and full-bandwidth seismic signal synthesis results from the corresponding iPhone signal. The first row shows the input (source) signal, the middle row is the transformed signal, and the bottom row shows the target (GT) signal from the Episensor.*

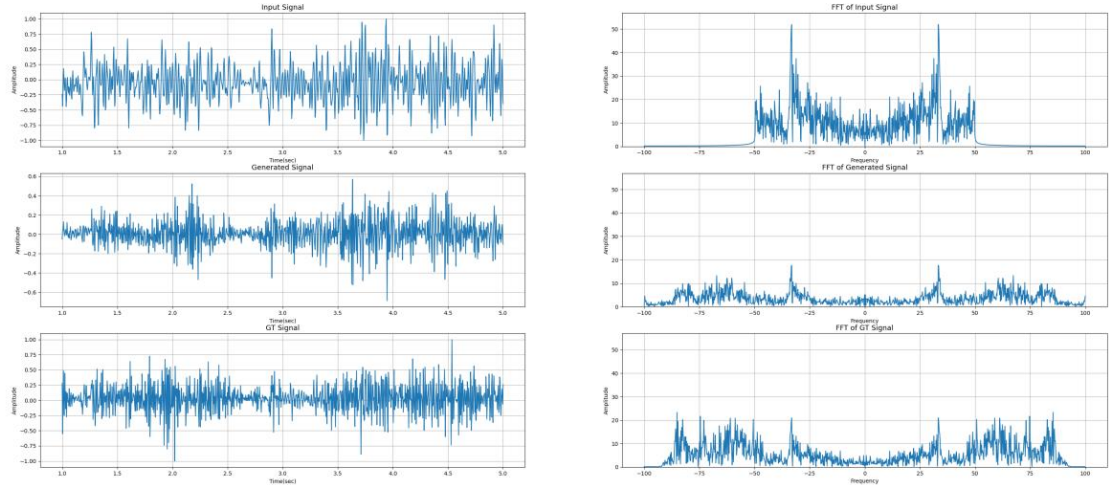

*Figure 10. High-quality and full-bandwidth seismic signal synthesis results from the corresponding iPhone signal. The first row shows the input (source) signal, the middle row is the transformed signal, and the bottom row shows the target (GT) signal from the Episensor.*

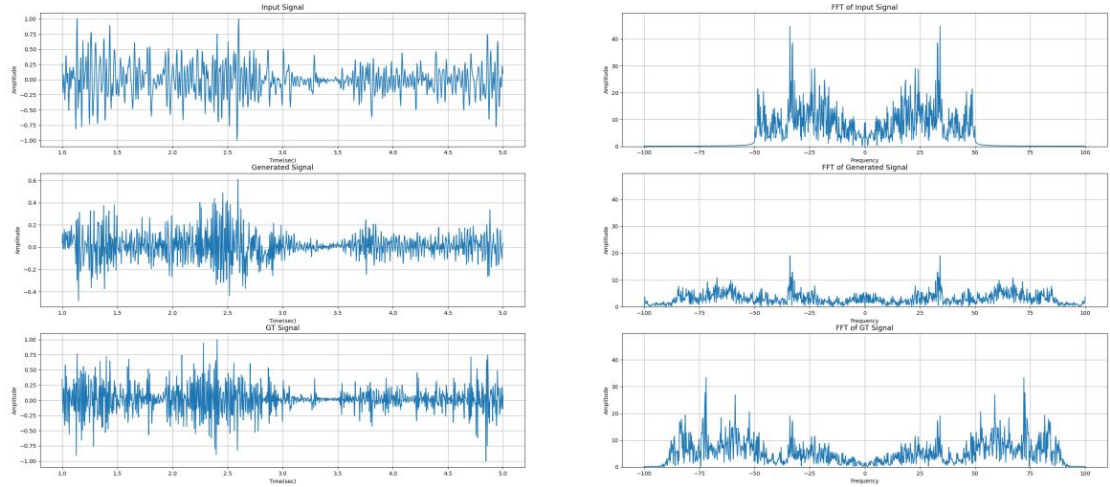

*Figure 11. High-quality and full-bandwidth seismic signal synthesis results from the corresponding iPhone signal. The first row shows the input (source) signal, the middle row is the transformed signal, and the bottom row shows the target (GT) signal from the Episensor.*

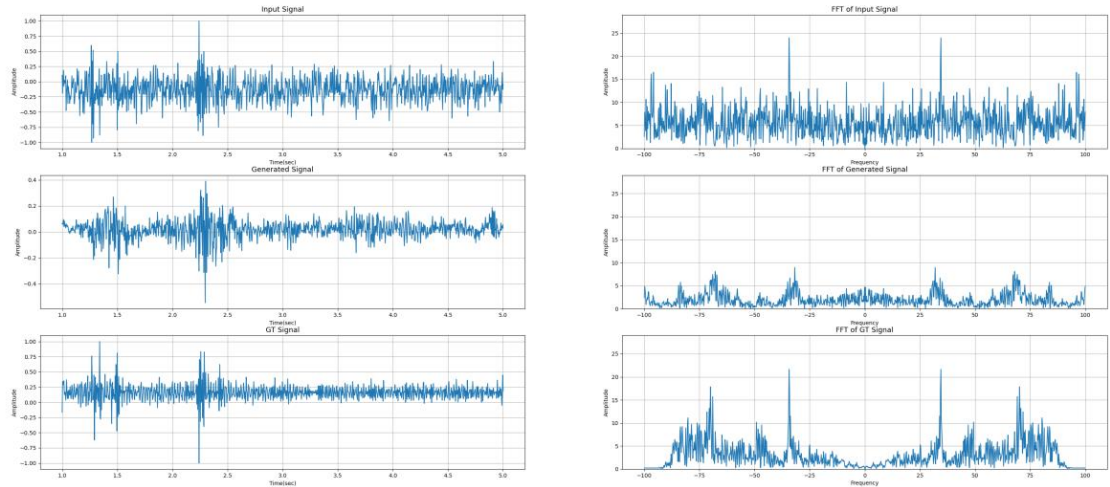

*Figure 12. High-quality and full-bandwidth seismic signal synthesis results from the corresponding Android signal. The first row shows the input (source) signal, the middle row is the transformed signal, and the bottom row shows the target (GT) signal from the Episensor.*

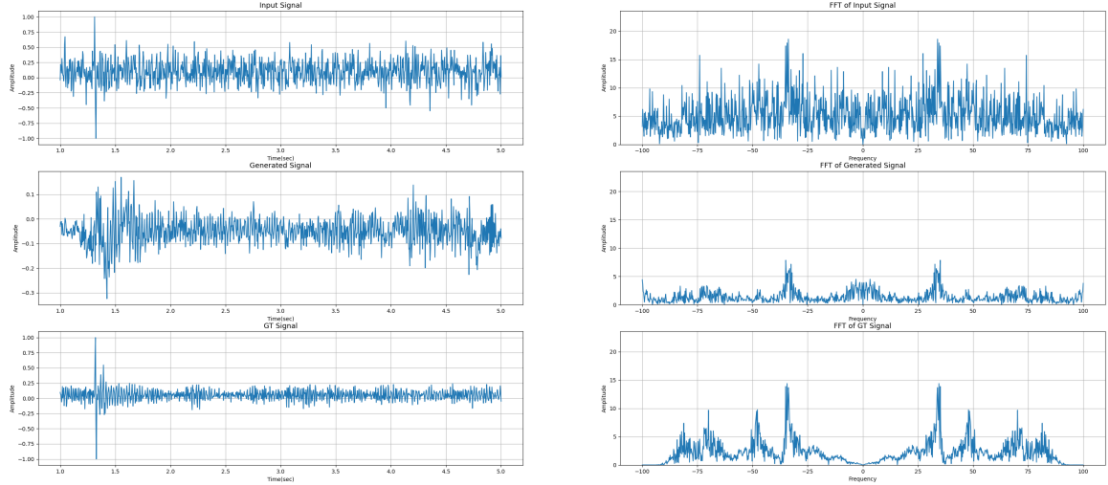

*Figure 13. High-quality and full-bandwidth seismic signal synthesis results from the corresponding Android signal. The first row shows the input (source) signal, the middle row is the transformed signal, and the bottom row shows the target (GT) signal from the Episensor.*

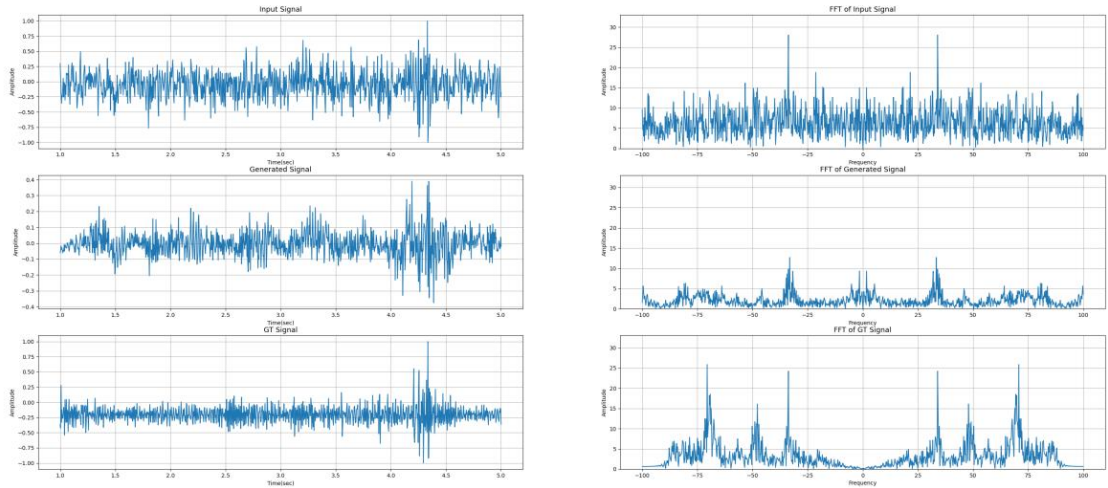

*Figure 14. High-quality and full-bandwidth seismic signal synthesis results from the corresponding Android signal. The first row shows the input (source) signal, the middle row is the transformed signal, and the bottom row shows the target (GT) signal from the Episensor.*